%% file: root.tex
\documentclass[letterpaper, 10 pt, conference]{Packages/ieeeconf}  
\IEEEoverridecommandlockouts                              
 
\usepackage{float}
\usepackage{times}
\usepackage{epsfig}
\usepackage{graphicx}
\usepackage{amsmath}
\usepackage{amssymb}
\usepackage[ruled]{algorithm}
\usepackage{algpseudocode}
\usepackage{ctable}
\usepackage[export]{adjustbox}
\usepackage{import}
\usepackage{pgf}
\usepackage{setspace}
\usepackage{url}
\usepackage{hyperref}
\usepackage{capt-of,etoolbox}
\usepackage{pifont} 

\makeatletter
\let\NAT@parse\undefined
\makeatother

\usepackage{Packages/sparo_acronyms}
\usepackage{Packages/sparo_math}
\usepackage{Packages/sparo_SIunits}
\usepackage{Packages/sparo_misc}
\usepackage{multirow}
\usepackage{indentfirst}

\usepackage[sort&compress,numbers]{natbib}

\usepackage{soul,color}
\usepackage{verbatim} 

\newcommand{\bbl}[1]{{\textcolor{blue}{#1}}}

\definecolor{bl}{HTML}{4285F4}
\newcommand{\bl}[1]{{\textcolor{bl}{#1}}}

\definecolor{rl}{HTML}{DB4437}
\newcommand{\rl}[1]{{\textcolor{rl}{#1}}}

\definecolor{gl}{HTML}{0F9D58}
\newcommand{\gl}[1]{{\textcolor{gl}{#1}}}

\DeclareMathOperator*{\argmin}{argmin}

\title{\LARGE \bf
	 SKiD-SLAM: Robust, Lightweight, and Distributed \\Multi-Robot LiDAR SLAM in Resource-Constrained Field Environments 
}

\author{Hogyun Kim${}^{1}$, Jiwon Choi${}^{1}$, Juwon Kim${}^{1}$, Geonmo Yang${}^{1}$, Dongjin Cho${}^{1}$, Hyungtae Lim${}^{2*}$, and Younggun Cho${}^{1*}$
\thanks{$^*$Corresponding authors. \hfill \break 
\indent ${}^{1}$H. Kim, ${}^{1}$J. Choi, ${}^{1}$J. Kim, ${}^{1}$G. Yang, ${}^{1}$D. Cho, and ${}^{1*}$Y. Cho are with the Dept. Electr. and Comput. Eng., Inha University, South Korea \texttt{(e-mail: [hg.kim, jiwon2, marimo117, ygm7422, d22g66]@inha.edu, yg.cho@inha.ac.kr)} \hfill \break 
\indent ${}^{2*}$H. Lim is with the Laboratory for Information \& Decision Systems (LIDS), Massachusetts Institute of Technology, Cambridge, MA 02139, USA. \texttt{(e-mail: shapelim@mit.edu)}}
}

\usepackage{fancyhdr}
\fancypagestyle{withfooter}{
  
  \fancyhf{}
  \fancyfoot[C]{Presented at the 2025 IEEE ICRA Workshop on Field Robotics}
}

\begin{document}
\maketitle
\thispagestyle{withfooter}
\pagestyle{withfooter}
\input{PaperWriting/abstract.tex}
\input{PaperWriting/introduction.tex} 
\input{PaperWriting/relatedwork.tex}

\input{PaperWriting/problem_formulation.tex} 
\input{PaperWriting/method.tex}
\input{PaperWriting/preliminary_experiments.tex}

\input{PaperWriting/field_experiments.tex}

\input{PaperWriting/conclusion.tex}

\scriptsize
\bibliographystyle{Packages/IEEEtranN} 
\bibliography{Packages/string-short, Packages/references}

\end{document}

%% file: PaperWriting/abstract.tex
\begin{abstract}
Distributed LiDAR SLAM is crucial for achieving efficient robot autonomy and improving the scalability of mapping.
However, two issues need to be considered when applying it in field environments: one is resource limitation, and the other is inter/intra-robot association.
The resource limitation issue arises when the data size exceeds the processing capacity of the network or memory, especially when utilizing communication systems or onboard computers in the field.
The inter/intra-robot association issue occurs due to the narrow convergence region of ICP under large viewpoint differences, triggering many false positive loops and ultimately resulting in an inconsistent global map for multi-robot systems.
To tackle these problems, we propose a distributed LiDAR SLAM framework designed for versatile field applications, called \textit{SKiD-SLAM}.
Extending our previous work that solely focused on lightweight place recognition and fast and robust global registration, we present a multi-robot mapping framework that focuses on robust and lightweight inter-robot loop closure in distributed LiDAR SLAM. 
Through various environmental experiments, we demonstrate that our method is more robust and lightweight compared to other state-of-the-art distributed SLAM approaches, overcoming resource limitation and inter/intra-robot association issues.
Also, we validated the field applicability of our approach through mapping experiments in real-world planetary emulation terrain and cave environments, which are in-house datasets.
Our code will be available at \textnormal{\url{https://sparolab.github.io/research/skid_slam/}}.
\end{abstract}

%% file: PaperWriting/introduction.tex
\section{Introduction}
Distributed \ac{LiDAR} \ac{SLAM} utilizing multi-robot platforms has been highlighted for achieving efficient robot autonomy and improving the scalability of mapping.
Particularly in field environments~(e.g., caves, planetary terrains, off-road areas, underground, aerial environments, and forests), multi-robot systems have been widely applied to achieve scientific exploration, planetary exploration \cite{arm2023scientific}, mine exploration \cite{mansouri2020deploying}, or industrial inspection.
To enable seamless collaboration, it is essential to utilize metadata (i.e., global descriptor) to recognize the inter-robot's location and leverage the full data (i.e., scan) to estimate relative poses between robots accurately.

Recently, various distributed \ac{LiDAR} \ac{SLAM} framework studies have demonstrated promising advances in mapping efficiency, drastically reducing processing time while enhancing scalability~\cite{huang2021disco, xie2021rdc, zhong2023dcl, lajoie2023swarm, he2025ldg}.
However, there are still two primary issues that existing methods have strived to solve, yet remain demanding in field applications.
One is the resource limitation, and the other is the inter/intra-robot association.

First, the resource limitation issue occurs when the data size exceeds the network or memory's processing capacity. 
Unlike well-established network infrastructures, field applications face communication constraints; therefore, it is essential to minimize the size of the global descriptor. 
Also, from a long-term autonomy perspective, the limited onboard memory of field robots necessitates lightweight global descriptors.
Second, existing distributed \ac{LiDAR} \ac{SLAM} mostly rely on \ac{ICP}-based local registration methods for inter-robot relative pose estimation, but these methods have narrow convergence regions and generally require a precise initial guess to achieve accurate results.
To maintain mapping accuracy, the narrow convergence regions need to be satisfied, which inherently limit the large pose discrepancy between robots.

\begin{figure*}[t]
	\centering
	\def\width{\textwidth}%
        {%
		\includegraphics[clip, trim= 0 0 30 0, width=\width]{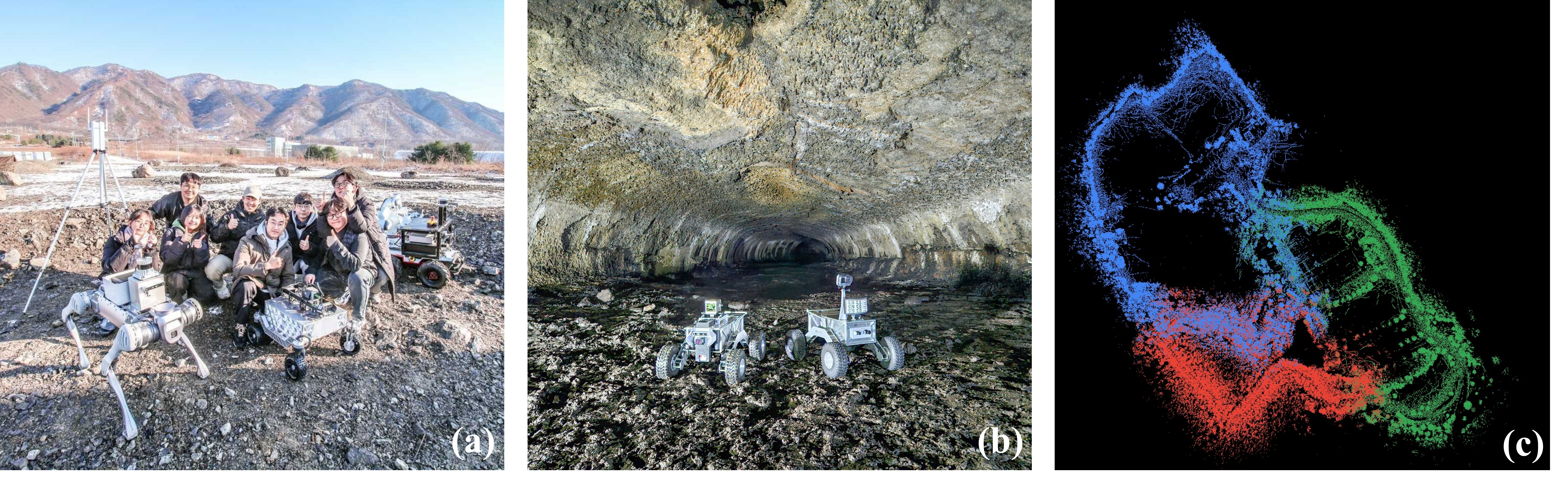}
	}
    \vspace{-0.8cm}
    \caption{(a) Team SPARO at the planetary emulation terrains with robots configured by sensors (e.g., LiDAR, IMU, RGB camera, etc), and wireless WiFi AP. 
             (b)~Custom multi-rover with sensors (e.g., LiDAR, IMU, thermal camera, etc) that was utilized for data acquisition in cave environments.
             (c)~Multi-robot mapping in the large-scale park (forest) dataset \cite{huang2021disco} using our \textit{SKiD-SLAM}.}
    \label{fig:main}
    \vspace{-0.5cm}
\end{figure*}

To address these two issues, we propose a \textit{SKiD-SLAM} for efficient collaboration in the field environments.
Extending our previous studies that solely focused on lightweight place recognition, \textit{SOLiD} \cite{kim2024narrowing} and fast and robust global registration, \textit{KISS-Matcher} \cite{lim2024kiss}, we address a multi-robot mapping framework in terms of the inter-robot loop closing in the distributed \ac{LiDAR} \ac{SLAM}. 
Our main contributions are as follows:
\begin{itemize}
    \item \textbf{Versatile Field Application:} We propose a novel distributed \ac{LiDAR} \ac{SLAM} framework called \textit{SKiD-SLAM}, which can be applied in various fields.
                                                Through experiments, we demonstrate that our method offers a robust solution for field robot applications in challenging environments, such as underground mine tunnels and off-road areas, where erroneous measurements may occur due to laser ray reflections.

    \item \textbf{Field Tests in Real-World:} As shown in \figref{fig:main}, we conducted various field tests in the cave and planetary emulation environments, which are in-house datasets. 
                                             During the field test, we also set up a Wi-Fi mesh network to conduct communication experiments.
                                                
                                                            
    \item \textbf{Lightweight and Accurate Inter/Intra-Loop Closure:} Building on our previous works, \textit{SOLiD} \cite{kim2024narrowing} and \textit{KISS-Matcher} \cite{lim2024kiss}, we propose a lightweight and accurate loop closure module designed for versatile field environment applications.
                                                                      Additionally, we demonstrate that our module achieves consistent performance under reverse or partially reverse loops.

\end{itemize}

%% file: PaperWriting/relatedwork.tex
\section{Related Works}
As reported by \citet{burgard2005coordinated}, the multi-robot system is often suggested to have several advantages over the single-robot system. 
First, the multi-robot system has the potential to perform a single task faster than a single robot.
Second, using multiple robots introduces inherent redundancy, enhancing the system's fault tolerance, which is expected to be more resilient than a single robot.
Finally, the overlapping information from the multi-robot helps compensate for sensor uncertainty, improving overall reliability in field environments, where erroneous measurements are prevalent.

Playing a key role in multi-robot LiDAR SLAM, inter/intra-robot loop closure is typically divided into three key components: (i) place recognition, (ii) registration, and (iii) consistency check.
First, place recognition finds links that are identified as identical (or similar) locations of inter/intra-robot using shared global descriptors. 
Second, once data retrieved via place recognition is identified as originating from the same location, registration estimates the relative pose of inter/intra-robot links based on shared scans. 
Finally, the consistency check verifies the relative pose of links between the two robots using scan data or each robot's local trajectory.
Based on these aspects, we analyze the existing multi-robot \ac{LiDAR} \ac{SLAM} frameworks.

Multi-robot LiDAR SLAM can be broadly classified into centralized LiDAR SLAM and distributed LiDAR SLAM.
Unlike centralized \ac{LiDAR} \ac{SLAM} \cite{kim2022lt, stathoulopoulos2024frame, kang2024unified}, which depends on a remote server to aggregate maps and compute global \ac{SLAM} for all robots, distributed \ac{LiDAR} \ac{SLAM} \cite{huang2021disco, xie2021rdc, zhong2023dcl, lajoie2023swarm, he2025ldg} depends on the robot's onboard computer, making it inherently constrained by limited resources.
Existing approaches utilize lightweight global descriptors to address this issue.
For instance, \citet{huang2021disco} proposed a DiSCo SLAM using Scan Context~\cite{kim2018scan}, which was designed for structured environments by dividing the surrounding space into bins and encoding each with the maximum elevation value.
\citet{xie2021rdc} proposed a RDC SLAM, which utilized DELIGHT~\cite{cop2018delight} for loop closure detection.
DELIGHT incorporates LiDAR intensity information into its descriptor, demonstrating its capability to operate in industrial parks with diverse characteristics, ranging from structured buildings to unstructured bushland.
\citet{zhong2023dcl} proposed DCL SLAM, which leverages LiDAR Iris~\cite{wang2020lidar}, representing LiDAR data as an image using Daugman’s rubber sheet model~\cite{john2009iris}.
This method was designed to achieve rotation-invariance using the Fourier Transform, enabling the detection of reverse or partial reverse loops, such as in intersections.
\citet{lajoie2023swarm} proposed a swarm SLAM using both LiDAR-based Scan Context \cite{kim2018scan} and vision-based place recognition, CosPlace \cite{berton2022rethinking}.
\citet{he2025ldg} proposed an LDG-CSLAM, which accumulates scans without an encoding process and utilizes submap-based NDT \cite{takeuchi20063}.

The aforementioned approaches maintain a lightweight data size.
However, in real-world field environments, where resources are even more constrained than networked infrastructures such as campuses or urban areas, there remains a persistent demand for a lighter global descriptor.

In field environments, even when inter/intra-robot systems traverse the same location, they may observe the environment from different viewpoints.
However, ICP-based local registration used in existing approaches~\cite{huang2021disco, zhong2023dcl}, with a narrow convergence region, can become trapped in local minima once a proper initial guess is not provided.
Namely, to preserve mapping accuracy, a narrow convergence region must be upheld, inherently constraining the large pose discrepancy between robots.

\begin{figure*}[t]
	\centering
	\def\width{0.99\textwidth}%
        {%
		\includegraphics[clip, trim= 0 40 0 100, width=\width]{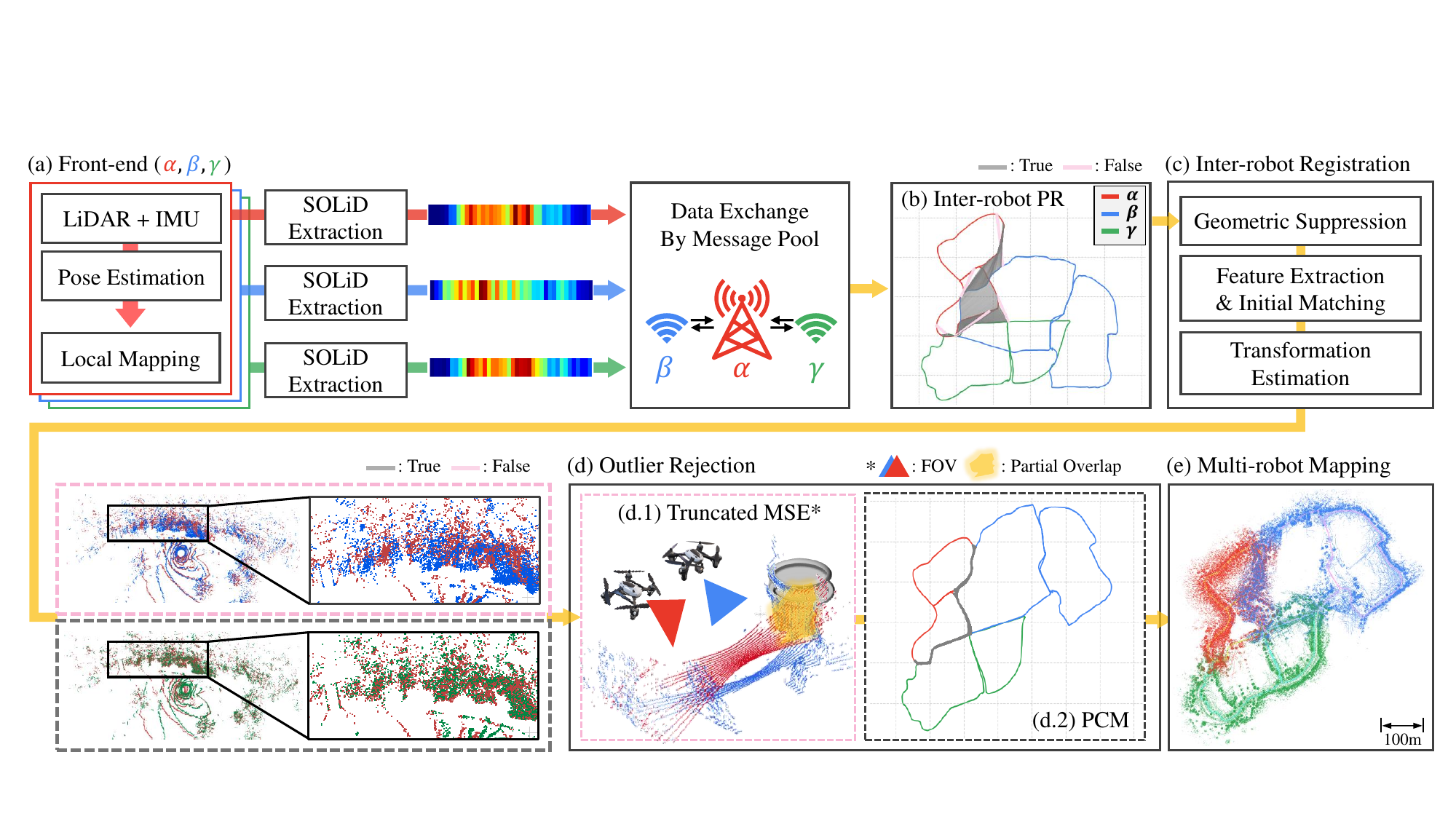}
	}
    \vspace{-0.35cm}
    \caption{Overview of our distributed LiDAR SLAM framework, called \textit{SKiD-SLAM}. 
    As an extension of our previous works, \textit{SOLiD} \cite{kim2024narrowing} and \textit{KISS-Matcher}~\cite{lim2024kiss}, \textit{SKiD-SLAM} is composed of five key components. Each robot is denoted as \textbf{\rl{$\alpha$}}, \textbf{\bl{$\beta$}}, and \textbf{\gl{$\gamma$}}, respectively.
    In the distributed case, robot $\alpha$ receives shared data from robots $\beta$ and $\gamma$.
    (a) Odometry estimation step using LiDAR-IMU fusion.
    (b) Inter-robot place recognition using robot $\alpha$'s \textit{SOLiD}s and associated robot's \textit{SOLiD}s.
    (c) Coarse-to-fine inter-robot registration using \textit{KISS-Matcher} and Small-GICP \cite{small_gicp} for coarse and find alignment, respectively.
    (d) Two-stage outlier rejection, where truncated mean square error (MSE) is used to verify the validity of each loop individually, and PCM \cite{mangelson2018pairwise} is used to verify the geometric consistency of loops across different trajectories.
    (e) Final distributed multi-robot mapping result centered on robot $\alpha$.
}
    \label{fig:pipeline}
    \vspace{-0.55cm}
\end{figure*}

In multi-robot systems, erroneous measurements can cause perceptual aliasing, leading to map distortions that degrade the accuracy and consistency of mapping.
This motivates the need for robust techniques capable of detecting and eliminating false positive links (or outliers), preventing severe distortions in multi-robot SLAM.
Therefore, these systems incorporate consistency checks to enhance the stability and reliability of mapping.
To this end, random sample consensus (RANSAC)~\cite{fischler1981random} is one of the renowned approaches that iteratively checks the geometric consistency of links, maintaining robust map merging.
Graduated non-convexity (GNC)~\cite{yang2020graduated} approximates a solution without an initial guess and then incrementally increases non-convexity at each iteration.
This approach enhances robustness against local minima and improves efficiency by reducing the need for extensive iterations, making it well-suited for distributed mapping systems \cite{tian2022kimera}.
Pairwise consistency maximization (PCM)~\cite{mangelson2018pairwise} evaluates pairwise consistency between each link of inter-robot loop closures to merge the multi-robot map robustly.
Each of these methods plays a critical role~\cite{huang2021disco} in ensuring robust multi-robot SLAM.
In particular, when applied in a two-stage process (e.g., RANSAC + PCM or GNC + PCM) \cite{zhong2023dcl, tian2022kimera}, these methods contribute to reducing incorrect inter-robot loop closures.

To address these challenges, we design the SKiD-SLAM, a robust, lightweight, distributed multi-robot LiDAR SLAM.
First, to enable minimal data exchange and rapid communication under the resource constraints of field robots, we employ the lightest global descriptor.
Second, to handle cases in which the large pose discrepancy between the viewpoints of two point clouds exceeds the convergence range of ICP, we employ a two-stage registration approach consisting of global and local registration.
Finally, to alleviate map distortion, we identify and then reject false positive links connecting inter/intra-robot measurements using a two-stage outlier rejection process at the pose graph level as a consistency check step.

%% file: PaperWriting/problem_formulation.tex
\section{Problem Formulation}
In this section, we formulate the objective function for our multi-robot LiDAR SLAM.
Let $\mathcal{X} = \{ \boldsymbol{x}_0, \cdots, \boldsymbol{x}_t \}$ be a set of keyframe poses in SE(3) from time 0 to time $t$.
The single-robot SLAM problem can then be formulated as a nonlinear least squares optimization:
\begin{equation}
\mathcal{X}^* = \argmin\limits_{\mathcal{X}} \sum \mathcal{F}(\mathcal{X}).
\label{equ:single_pose}
\end{equation}
The single robot pose graph $\mathcal{F}$ is defined as:
\begin{equation}
\begin{aligned}
    \mathcal{F}(\mathcal{X}) = &  \sum\limits_{\langle t, t+1 \rangle \in \mathcal{C^\text{odom}}}\left\|f\left(\boldsymbol{x}_t, \boldsymbol{x}_{t+1}\right)
                    - \hat{\boldsymbol{z}}_{t, t+1} \right\|_{\Omega_{t, t+1}}^2 \\
                  & + \sum_{\langle i, j \rangle \in \mathcal{C_\text{intra}^\text{loop}}}                
                      \left(\left\|f\left(\boldsymbol{x}_i, \boldsymbol{x}_j\right)
                    - \kappa(\boldsymbol{z}_i,\boldsymbol{z}_j) \right\|_{\Omega_{i, j}}^2 \right),
\end{aligned}
\label{equ:single_pgo}
\end{equation}
where $f(\cdot, \cdot)$ is a relative pose function, which estimates the motion between two frames, $\hat{\boldsymbol{z}}$ is estimated odometry, $\mathcal{C^\text{odom}}$ is an odometry constraint set between two consecutive $t$-th and $t+1$-th poses of a robot, $\mathcal{C^\text{loop}_{\text{intra}}}$ is a loop constraint set between two non-consecutive $i$-th and $j$-th poses of a robot (i.e., $i+1 \neq j$), and $\kappa(\cdot, \cdot)$ is a relative pose estimator between non-consecutive frames.
By minimizing the Mahalanobis distance ($|| \cdot ||^2_{\Omega}$) weighted by the information matrices $\Omega_{t, t+1}$ and $\Omega_{i,j}$, we optimize the pose set to best align the estimated and observed transformations.
Note that $\mathcal{C^\text{odom}}$ and $\mathcal{C^\text{loop}_{\text{intra}}}$ are disjoint, so $\mathcal{C^\text{odom}}$ $\cap$ $\mathcal{C^\text{loop}_{\text{intra}}}$ $= \varnothing$.

Next, we extend \equref{equ:single_pgo}, which applies only to a single robot, to multi-robot systems.
Let us assume that we deploy $N$ robots: $\mathcal{N} = \{ 1, 2,\cdot\cdot\cdot, n,\cdot\cdot\cdot, N \},$ then $\mathcal{X}_n$ represents the poses of the $n$-th robot and $\mathcal{X}_{\text{total}} = \{ \mathcal{X}_n|n \in \mathcal{N} \}$ is a set of all $N$ robots' keyframe poses.
Then, we can solve the multi-robot SLAM problem as follows:
\begin{equation}
    \mathcal{X}^*_{\text{total}} = \underset{ \mathcal{X}_{\text{total}}}{\arg\min} (\mathcal{F}_{\text{intra}} + \mathcal{F}_{\text{inter}}),
\label{equ:multi_pose}
\end{equation}
where $\mathcal{F}_{\text{intra}}$ represents the inter-robot constraints for each robot and is defined as follows:
\begin{equation}
    \mathcal{F}_{\text{intra}} = \underset{n \in \mathcal{N}}\sum\sum\limits_{\mathcal{C}_n} \mathcal{F}(\mathcal{X}_n),
\label{equ:multi_intra}
\end{equation}
where $\mathcal{C}_n$ denotes the set of all constraints for the $n$-th robot, consisting of $\mathcal{C}^{\text{odom}}$ and $\mathcal{C}^{\text{loop}}_{\text{intra}}$ (i.e., $\mathcal{C}_n = \mathcal{C}^{\text{odom}}\cup\mathcal{C}^{\text{loop}}_{\text{intra}}$).

\noindent $\mathcal{F}_{\text{inter}}$ denotes the inter-robot loop closure constraints, which represent correspondences across robots as follows:
\begin{equation}
    \mathcal{F}_{\text{inter}} =  \sum\limits_{n, m \in \mathcal{N}}
                                  \sum\limits_{\langle k, l \rangle \in \mathcal{C_\text{inter}^\text{loop}}}                
                                  \left(\left\|f\left(\boldsymbol{x}_k, \boldsymbol{x}_l\right)
                                - \kappa(\boldsymbol{z}_k,\boldsymbol{z}_l) \right\|_{\Omega_{k, l}}^2 \right),
\label{equ:multi_inter}
\end{equation}
where $\mathcal{C^\text{loop}_{\text{inter}}}$ is an inter-robot loop constraint set between two different robots' $k$-th and $l$-th poses for the $n$-th and $m$-th robots (i.e., $\boldsymbol{x}_k, \boldsymbol{x}_l$) with their observed relative pose (i.e., $\boldsymbol{z}_k, \boldsymbol{z}_l$), $\Omega_{k, l}$ is the information matrix, and $m$ represents the $m$-th robot among the $N$ robots, excluding the $n$-th robot (i.e., $m \in \mathcal{N} \backslash \{n\}$). 

%% file: PaperWriting/method.tex
\section{SKiD SLAM: Robust, Lightweight, and Distributed Multi-Robot LiDAR SLAM}
Our SKiD-SLAM is structured as shown in \figref{fig:pipeline}.
The framework consists of five main components: (i) front-end comprising odometry estimation, SOLiD extraction, and local mapping, (ii) inter/intra-robot place recognition using the SOLiD shared through the message pool, (iii) inter/intra-robot registration to estimate the relative pose between the robots, (iv) outlier rejection to filter out false links between the robots, and (v)~unified global mapping of multi-robot.
Since both inter- and intra-robot place recognition and registration share the same module, this section explains the procedure from an inter-robot perspective.

\subsection{System Overview}
As the number of robots increases, centralized SLAM becomes computationally demanding on the central server~\cite{lajoie2022towards}. 
In particular, carrying a high-performance server in the field environments is highly inefficient. 
To address this, we adopt a distributed SLAM approach.
In terms of distributed SLAM, we reformulate \equref{equ:multi_intra} and \equref{equ:multi_inter} into an objective function for the $\alpha$ robot as follows:
\begin{align}
     & \mathcal{X}_{\text{dis},\alpha}^{*}  = \underset{\mathcal{X}_{\text{dis},\alpha}}{\arg\min}
    \bigg(\underset{\alpha \in \mathcal{N}_{\alpha}}\sum\sum\limits_{\mathcal{C}_{\text{dis},\alpha}} \mathcal{F}(\mathcal{X}_{\alpha}) \; + \notag \\
       &  \sum\limits_{\alpha, \beta \in \mathcal{N}_\alpha}
                                  \sum\limits_{\langle c, q \rangle \in \mathcal{C_\text{dis,inter}^\text{loop}}}                
                                  \left(\left\|f\left(\boldsymbol{x}_c, \boldsymbol{x}_q\right)
                                - \kappa(\boldsymbol{z}_c,\boldsymbol{z}_q) \right\|_{\Omega_{c, q}}^2 \right)
    \bigg),
\label{equ:dis_pgo}
\end{align}


where $\mathcal{N_\alpha} = \{\alpha, \beta, \cdots \}$ is the set of indices of robots associated with the $\alpha$ robot (e.g., robots that can have links connected to robot $\alpha$ within 30\,m), $\mathcal{C_{\text{dis},\alpha}} \subset \mathcal{C}_n$ is the union of partial odometry constraints, $\mathcal{C}^\text{odom}_{\text{dis}} \subset \mathcal{C^\text{odom}}$, and partial intra-robot loop constraints, $\mathcal{C_{\text{dis,intra}}^\text{loop}} \subset \mathcal{C_\text{intra}^\text{loop}}$, related to robot $\alpha$, $\mathcal{C^\text{loop}_{\text{dis,inter}}} \subset \mathcal{C^\text{loop}_{\text{inter}}}$ is a inter-robot loop constraints set between robot $\alpha$ and $\beta$ (i.e., $\mathcal{N}_\alpha \backslash \{ \alpha \}$) robot $c$-th and $q$-th poses for the $\alpha$ and $\beta$ robots (i.e., $\boldsymbol{x}_c, \boldsymbol{x}_q$) with their observed relative pose (i.e., $\boldsymbol{z}_c, \boldsymbol{z}_q$), and $\Omega_{c, q}$ is the information matrix.
$\mathcal{X}_{\text{dis},\alpha}$, which contains all poses related to robot $\alpha$, is defined as follows:
\begin{equation}
    \mathcal{X}_{\text{dis},\alpha} = \mathcal{X}_{\alpha} \cup 
    \left\{ \boldsymbol{x}_q \; \middle|
    \begin{aligned}
        & \; \boldsymbol{x}_q \in \mathcal{X}_{\beta}, \langle c, q \rangle \in \mathcal{C^\text{loop}_{\text{dis,inter}}}, \\
        & \; \forall \beta \in \mathcal{N_\alpha}, \beta \neq \alpha
    \end{aligned}
    \right\}.
\end{equation}

Unlike \equref{equ:multi_pose}, which optimizes over all $N$ robots, we operate the efficient objective function \equref{equ:dis_pgo}, which optimizes only for robot $\alpha$ and its associated robots.

\subsection{Inter-robot Place Recognition}
Inter-robot place recognition is the problem of determining $\mathcal{C^\text{loop}_{\text{dis,inter}}}$ in \equref{equ:dis_pgo} whether two robots $\alpha$ and $\beta$ have observed the same place (e.g., the $c$-th place in robot $\alpha$ and $q$-th place in robot $\beta$).
To solve this problem, we utilize the SOLiD, which is a lightweight, robust, and rotation-invariant descriptor.
We achieved this by embedding their respective point cloud, $\mathcal{P}_c^\alpha$ and $\mathcal{P}_q^\beta$, into global descriptors ($\mathbf{D}_{c}^\alpha = \omega(\mathcal{P}_{c}^\alpha) \in \mathbb{R}^e$, $\mathbf{D}_q^\beta = \omega(\mathcal{P}_q^\beta)\in \mathbb{R}^e$), where $\omega(\cdot)$ is an embedding function and $e$ is an embedding dimension.
Then, we measure the distance $d$ between the descriptors, and verify the conditions $\tau_\text{dist}$ as: 
\begin{equation}
   d = \phi (\mathbf{D}_c^{\alpha}, \mathbf{D}_q^\beta), \quad s.t. \;\; d < \tau_\text{dist},
\label{equ:solid}
\end{equation}
where $\phi(\cdot, \cdot)$ is a Euclidean distance-measuring function.

\begin{figure}[t]
	\centering
	\def\width{0.48\textwidth}%
        {%
		\includegraphics[clip, trim= 0 60 0 80, width=\width]{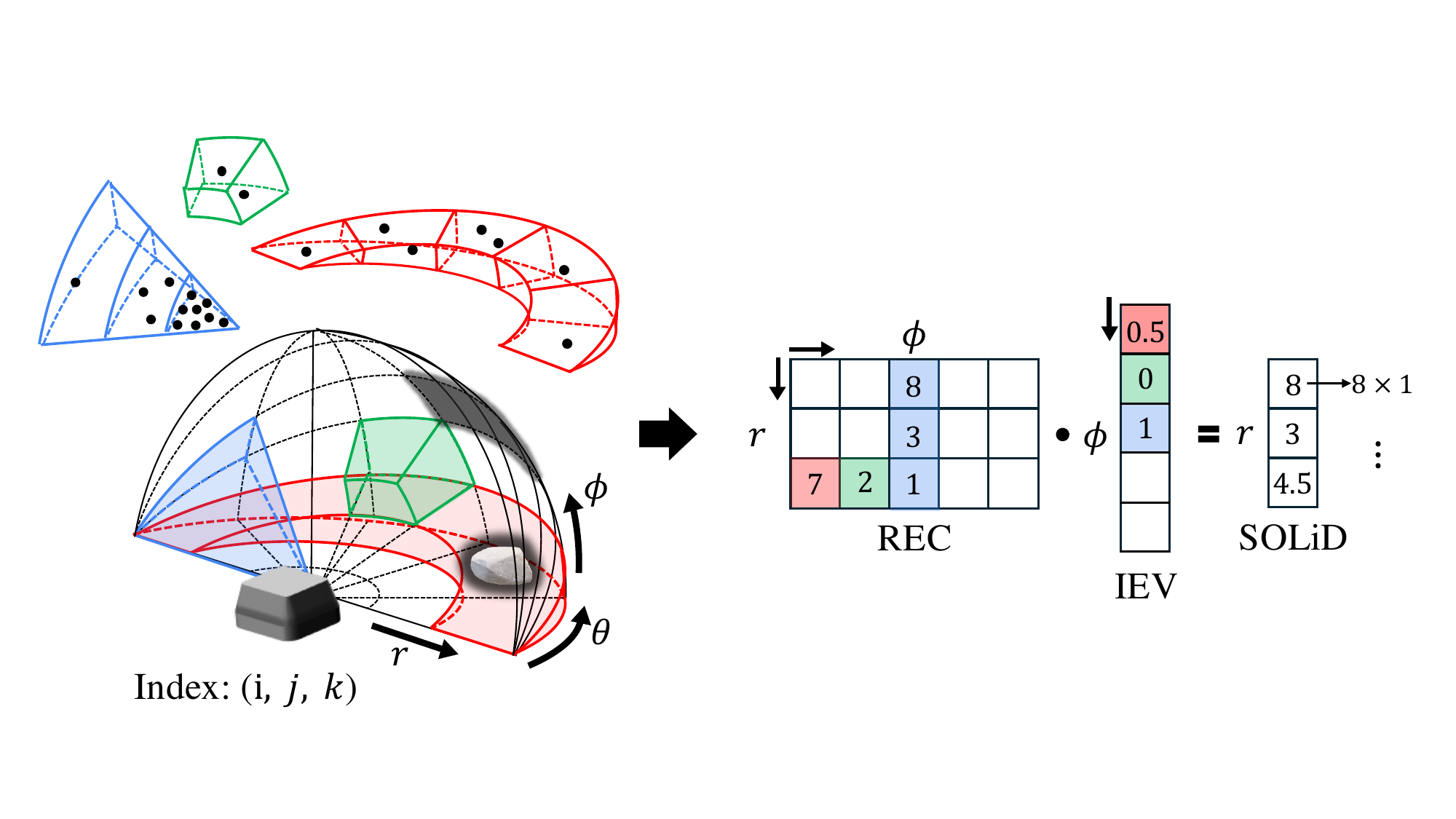}
	}
    \vspace{-0.7cm}
    \caption{SOLiD constructs a 2D bin representation called radial-elevational points counter (REC), consisting of range and elevation, and defines the bin's representative value based on the number of points within each bin.
            Next, an implicit elevation vector (IEV) is generated by performing summation along the range direction of REC, followed by min-max normalization. 
            The final SOLiD is then obtained by computing the dot product between REC and IEV.
            For example, assuming the green points in \figref{fig:solid} represent occlusion or erroneous measurements caused by reflections of laser rays, directly using them (e.g., REC) could lead to perceptual aliasing.
            However, our approach leverages the dot product operation, which propagates lower values as low and higher values as high, effectively handling challenging field environments.
            } 
    \label{fig:solid}
    \vspace{-0.6cm}
\end{figure}

Using SOLiD \cite{kim2024narrowing} offers three key advantages over existing methods in addressing the \equref{equ:solid}.
First, SOLiD is applicable across a range of FOVs from 60$^\circ$ to 360$^\circ$ and remains effective even under occlusion conditions.
Second, SOLiD maintains robustness compared to relying on geometric or intensity features in field environments.
For instance, Scan Context \cite{kim2018scan}, which uses the geometric features (i.e., elevation value for each bin) as the representative value in the same ($r$, $\theta$) space, may overlook key features such as the rocks in \figref{fig:solid}.
In contrast, our method maintains robust performance even in erroneous measurements.
Additionally, in environments such as caves where the intensity values of individual points exhibit little variation \cite{novakova2022correcting}, extracting remarkable intensity features becomes challenging.

As in \cite{kim2018scan}, we also design the distance-measuring function $\phi(\cdot, \cdot)$ in \equref{equ:solid}, as a KD-Tree, enabling faster correspondence retrieval.
In sum, the pair $ \langle c, q \rangle $ that satisfies \equref{equ:solid} through SOLiD is considered an element of the set $\mathcal{C^\text{loop}_{\text{dis,inter}}}$.

\subsection{Inter-robot Registration}
Inter-robot registration utilizing $\kappa(\cdot, \cdot)$ in \equref{equ:dis_pgo} estimates the relative pose between $\mathcal{P}_c^\alpha$ and $\mathcal{P}_q^\beta$ at the same places of two robots, $\alpha$ and $\beta$ in $\mathcal{C^\text{loop}_{\text{dis,inter}}}$.
To estimate the relative pose, we utilize the KISS-Matcher, which is a robust global registration method.
Let us assume that the $ \langle c, q \rangle $ pair obtained through matching consists of the 3D points $\mathbf{p}_c^\alpha \in \mathcal{P}_c^\alpha$ and the 3D points $\mathbf{p}_q^\beta \in \mathcal{P}_q^\beta$.
Then, the $\kappa(\mathcal{P}_c^\alpha, \mathcal{P}_q^\beta)$ can be expressed as follows:
\begin{equation}
    \hat{\mathbf{R}}, \hat{\mathbf{t}} = \argmin\limits_{\mathbf{R}, \mathbf{t}}\sum\limits_{\langle c, q \rangle \in \mathcal{A} \backslash \hat{\mathcal{O}}}  \rho \bigg(\left\|(\mathbf{p}_q^\beta - \mathbf{R}\mathbf{p}_c^\alpha) - \mathbf{t} \right\|_2 \bigg),
\label{equ:kiss}
\end{equation}
where $\mathbf{R} \in \text{SO(3)}$ is the relative rotation matrix, $\mathbf{t} \in \mathbb{R}^3$ is the translation vector, $\rho(\cdot)$ is a robust kernel for suppressing large errors caused by false correspondences, $||\cdot||_2$ is a Euclidean distance, $\mathcal{A}$ is an initial correspondence set, and $\mathcal{O}$ denotes a set of estimated false correspondences.

As reported by \citet{lim2024kiss}, in the presence of a large pose discrepancy, ICP-based local registration may fail to converge to the optimal solution as the viewpoint difference becomes larger.
However, once a reliable initial guess for the relative pose is available as a coarse alignment, the local registration can reliably converge to the global optimum as a fine alignment.
For example, \citet{kim2018scan} provides the place recognition result, $d$ in \equref{equ:solid} with initial guess, thereby demonstrating effective use of ICP in reverse loops, such as those encountered at intersections.
Similarly, we employ our previous work, KISS-Matcher, to estimate the accurate initial guess.

Using KISS-Matcher, we gain two key benefits.
First, we reduce false correspondences that arise from repeating patterns in the environment \cite{yang2020teaser} by applying geometric suppression.
This allows us to perform a sanity check on the reliability of the relative pose by verifying the cardinality of $\mathcal{A} \backslash \hat{\mathcal{O}}$.
Second, it introduces Faster-PFH, a more efficient alternative to Fast Point Feature Histograms (FPFH) \cite{rusu2009fast}, which delivers similar performance while enabling robust registration.
This enables us to formulate the objective function in \equref{equ:kiss} without false correspondences, ultimately deriving $\mathbf{R}$ and $\mathbf{t}$ to be converged to the global optimal through fine alignment, Small-GICP \cite{small_gicp}.

\subsection{Inter-robot Outlier Rejection}
Outlier rejection is introduced to prevent false positive loop closures in place recognition, which could otherwise disrupt multi-robot mapping.
Existing studies mostly have addressed this by checking consistency through each robot's local trajectory~\cite{mangelson2018pairwise}.
However, erroneous measurements are common in field environments.
Therefore, we introduce a two-stage outlier rejection that not only checks consistency but also evaluates inliers and fitness scores between matched point clouds ($\mathcal{P}_{c}^\alpha$, $\hat{\mathcal{P}}_q^\beta = \mathbf{R}\mathcal{P}_q^\beta + \mathbf{t}$) at the measurement level similar to \cite{zhong2023dcl, tian2022kimera}.

\subsubsection{Truncated MSE} We reject outliers based on the fitness score, $s$, a metric that evaluates the alignment quality between matched point clouds from $\mathcal{P}_{c}^\alpha$ and $\hat{\mathcal{P}}_q^\beta$ called geometric verification.
However, when the robots move in different directions, reverse or partially reverse overlapping occurs. 
In this case, the fitness score increases quadratically.
Thus, we apply the truncated mean squared error (MSE) to focus on the overlapping region as much as possible when computing the fitness score:
\begin{equation}
    s = \frac{1}{\mathbf{N}_{\text{inliers}}}\sum\limits_{i=1}^{\mathbf{N}_{\text{inliers}}} \mathbb{I} \;(||\mathbf{p}_{c_i}^\alpha - \hat{\mathbf{p}}_{q_i}^\beta||_2 \; <  \; \tau_\text{MSE} ),
\end{equation}
where $||\cdot||_2$ is a Euclidean distance between matching pair of points, ($\mathbf{p}_{c_i}^\alpha$, $\mathbf{p}_{q_i}^\alpha$), $\mathbf{N}_{\text{inliers}}$ is the number of inlier matching pairs, $\mathbb{I}(\cdot)$ is an indicator function that returns distance if the given condition is true, and $\tau_\text{MSE}$ is a user-defined maximum correspondence distance.
This function not only aids in robust multi-robot SLAM but also enables robots' flexible responses to the large pose discrepancy between robots.

\subsubsection{PCM}
Let $\{\boldsymbol{x}_{c_1}, \boldsymbol{x}_{c_2}\} \in \mathcal{X}_\alpha$, $\{\boldsymbol{x}_{q_1}, \boldsymbol{x}_{q_2} \} \in \mathcal{X}_\beta$ be the $c_1$-th, $c_2$-th poses in robot $\alpha$ and $q_1$-th, $q_2$-th poses in robot $\beta$.
Then, we can derive intra-robot relative measurements, $\boldsymbol{z}_{c_1 c_2}$, $\boldsymbol{z}_{q_2 q_1}$, utilizing the $\mathcal{C}_{\text{dis}}^{\text{odom}}$ of \equref{equ:dis_pgo} and inter-robot measurements, $\boldsymbol{z}_{c_1 q_1}$, $\boldsymbol{z}_{c_2 q_2}$, leveraging the $\mathbf{R}$ and $\mathbf{t}$ in \equref{equ:kiss}.
Next, we apply PCM~\cite{mangelson2018pairwise} to enhance the system's stability by rejecting loop closure outliers via consistency, $p$, as follows:
\begin{equation}
    p = \left\| (\ominus \boldsymbol{z}_{c_1 q_1}) \oplus \boldsymbol{z}_{c_1 c_2} \oplus \boldsymbol{z}_{c_2 q_2} \oplus \boldsymbol{z}_{q_2 q_1} \right\|_\Omega < \tau_{\text{PCM}},
\end{equation}

where $\oplus$ and $\ominus$ denote pose composition and inversion using the notation from \cite{smith1990estimating} and $\tau_{\text{PCM}}$ is the user-defined threshold.  
After performing multiple pairwise consistency checks by solving a maximum clique problem.
The selected loop closures from this set are accepted and updated as final loops for $\mathcal{C_\text{dis,inter}^\text{loop}}$, then passed to the back-end distributed PGO for further joint optimization in \equref{equ:dis_pgo}, contributing to robust and reliable mapping in the field environment.


%% file: PaperWriting/preliminary_experiments.tex

\section{Preliminary Evaluation}
In this section, we conducted a preliminary assessment using publicly available field datasets, laying the foundation for planetary emulation terrains and cave exploration, which are our field tests of real-world as shown in Section \uppercase\expandafter{\romannumeral6}.
For easy comprehension, we differentiated experiment results: \bbl{\textbf{first rank}} in blue bold.

\begin{figure}[t]
	\centering
    \def\width{0.45\textwidth}%
        {%
		\includegraphics[clip, trim= 0 0 30 0, width=\width]{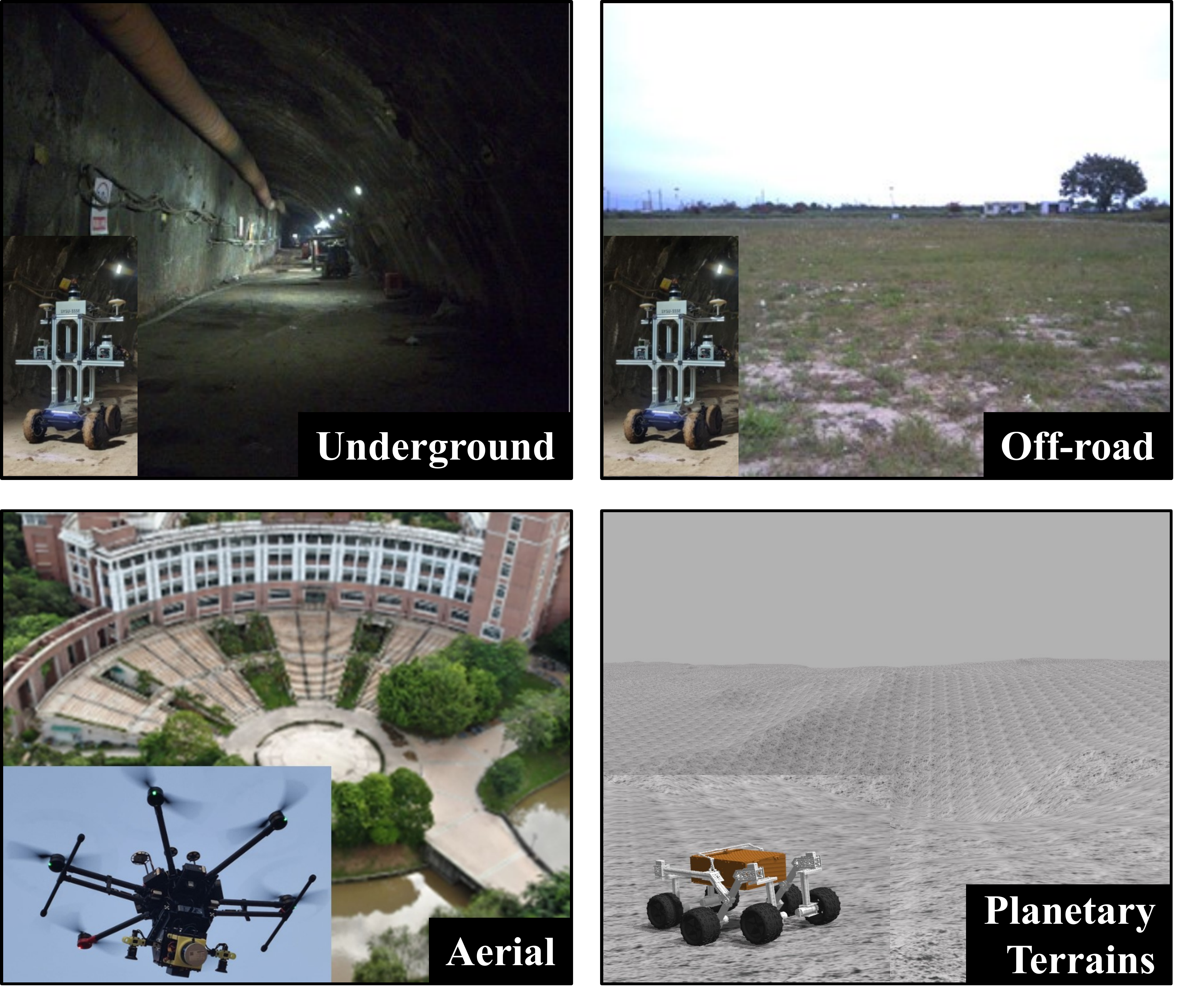}
	}
    \vspace{-0.4cm}
    \caption{Public datasets for our preliminary evaluation.
    Underground mine tunnels and off-road areas are sequences from the GEODE dataset \cite{chen2024heterogeneous}.
    Aerial environments and the moon are sequences from the GRACO dataset \cite{zhu2023graco} and our simulator's custom sequence, respectively.
    }
    \vspace{-0.1cm}
    \label{fig:dataset}
\end{figure}

\begin{table}[t]
\caption{Comparison of Total Descriptor Memory Usage for Mapping the Park Scene}
\renewcommand{\arraystretch}{1.2}
\centering\resizebox{0.48\textwidth}{!}{\tiny
\begin{tabular}{l|l|ccc}
\toprule \hline
                                & \textbf{Sharing}                                & \textbf{Scan Context}  & \textbf{LiDAR Iris}  & \textbf{Ours} \\ \hline
 \multirow{2}{*}{\textbf{PARK}} & \multirow{1}{*}{$\beta \rightarrow \alpha$}     & 122.546$\,$MB          & 5.996$\,$GB          & \bbl{\textbf{5.481}$\,$\textbf{MB}}     \\
                                & \multirow{1}{*}{$\gamma \rightarrow \alpha$}    & 156.429$\,$MB          & 7.654$\,$GB          & \bbl{\textbf{6.997}$\,$\textbf{MB}}     \\ \hline
\bottomrule
\end{tabular}}
\label{tab:desc}
\vspace{-0.6cm}
\end{table}

\begin{figure*}[t]
	\centering
	\def\width{0.97\textwidth}%
        {%
		\includegraphics[clip, trim= 0 90 0 0, width=\width]{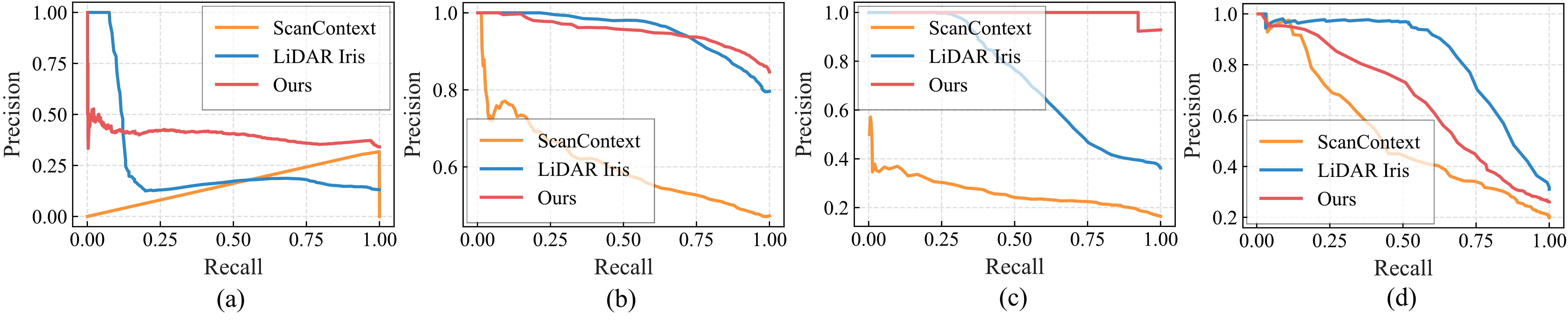}
	}
    \vspace{-0.45cm}
    \caption{(Left$\rightarrow$Right): PR curve in the aerial environments, underground mine tunnel, off-road areas, and planetary terrains.}
    \label{fig:pr_curve}
    \vspace{-0.2cm}
\end{figure*}

\begin{figure*}[h!]
	\centering
	\def\width{0.97\textwidth}%
        {%
		\includegraphics[clip, trim= 0 90 0 0, width=\width]{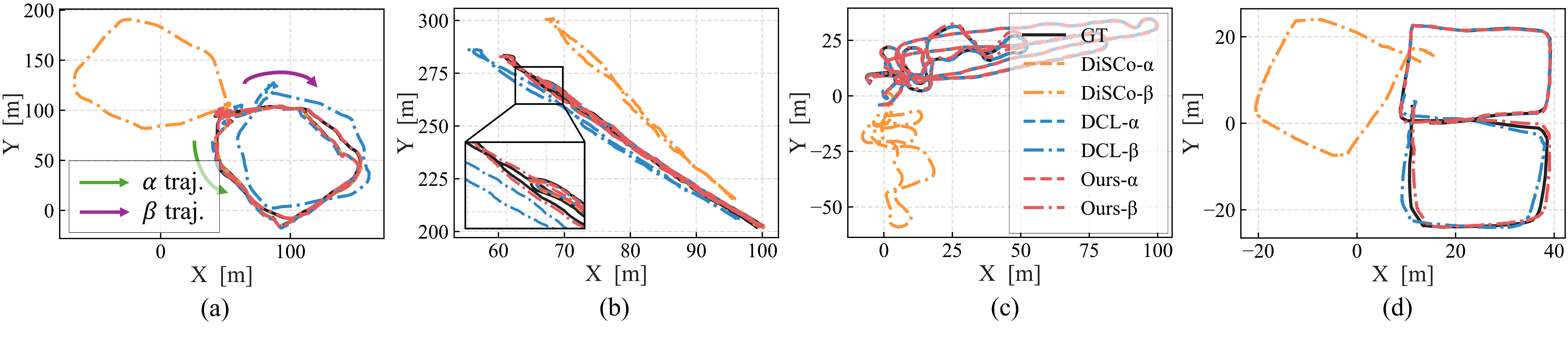}
	}
    \vspace{-0.45cm}
    \caption{(Left$\rightarrow$Right): Trajectories evaluation in the aerial environments, underground mine tunnel, off-road areas, and planetary terrains.}
    \label{fig:trajectories}
    \vspace{-0.1cm}
\end{figure*}

\begin{table*}[h!]
    \caption{Comparison of absolute translation error (ATE) and absolute rotation error (ARE) using Distributed LiDAR SLAM approaches}
    \renewcommand{\arraystretch}{1.4}
    \centering\resizebox{0.97\textwidth}{!}{
    \begin{tabular}{l|cccc|cccc|cccc|cccc}
     \toprule \hline
   \textbf{Datasets}    & \multicolumn{4}{c|}{\textbf{Aerial Environments}}      & \multicolumn{4}{c|}{\textbf{Underground}}   & \multicolumn{4}{c|}{\textbf{Off-road areas}}   & \multicolumn{4}{c}{\textbf{Planetary Terrains}} \\ \hline
    \textbf{Evaluation} & \multicolumn{2}{c|}{\textbf{ATE [m]}}     & \multicolumn{2}{c|}{\textbf{ARE [deg]}} & \multicolumn{2}{c|}{\textbf{ATE [m]}}    & \multicolumn{2}{c|}{\textbf{ARE [deg]}}
                        & \multicolumn{2}{c|}{\textbf{ATE [m]}}     & \multicolumn{2}{c|}{\textbf{ARE [deg]}} & \multicolumn{2}{c|}{\textbf{ATE [m]}}    & \multicolumn{2}{c}{\textbf{ARE [deg]}} \\ \hline
    \textbf{Robot Name} & robot $\alpha$                           & \multicolumn{1}{c|}{robot $\beta$}       & robot $\alpha$                          & robot $\beta$     
                        & robot $\alpha$                           & \multicolumn{1}{c|}{robot $\beta$}       & robot $\alpha$                          & robot $\beta$      
                        & robot $\alpha$                           & \multicolumn{1}{c|}{robot $\beta$}       & robot $\alpha$                          & robot $\beta$     
                        & robot $\alpha$                           & \multicolumn{1}{c|}{robot $\beta$}       & robot $\alpha$                          & robot $\beta$      \\ \hline
    DiSCo-SLAM \cite{huang2021disco} & \bbl{\textbf{0.892}}         & \multicolumn{1}{c|}{131.106}             & 80.404                                  & 164.113 
                        & 0.752                                    & \multicolumn{1}{c|}{16.938}              & N/A*                                    & N/A*  
                        & \bbl{\textbf{0.143}}                      & \multicolumn{1}{c|}{46.583}              & N/A*                                    & N/A*     
                        & 0.137                                    & \multicolumn{1}{c|}{40.924}              & \bbl{\textbf{0.316}}                     & 2.380           \\ \hline
    DCL-SLAM \cite{zhong2023dcl}     & 7.953                       & \multicolumn{1}{c|}{39.600}              & 6.405                                   & 29.886
                        & 0.318                                    & \multicolumn{1}{c|}{5.551}               & N/A*                                    & N/A*      
                        & 0.287                                    & \multicolumn{1}{c|}{\bbl{\textbf{0.313}}} & N/A*                                    & N/A*
                        & \bbl{\textbf{0.120}}                      & \multicolumn{1}{c|}{1.851}               & 0.364                                   & 5.674           \\ \hline
    SKiD-SLAM (Ours)    & 1.191                                    & \multicolumn{1}{c|}{\bbl{\textbf{1.058}}} & \bbl{\textbf{2.167}}                     & \bbl{\textbf{1.716}} 
                        & \bbl{\textbf{0.339}}                      & \multicolumn{1}{c|}{\bbl{\textbf{0.403}}} & N/A*                                    & N/A*   
                        & 0.300                                    & \multicolumn{1}{c|}{0.925}               & N/A*                                    & N/A*   
                        & 0.198                                    & \multicolumn{1}{c|}{\bbl{\textbf{1.046}}} & 0.546                                   & \bbl{\textbf{0.080}}   \\ \hline \bottomrule
                        \multicolumn{17}{r}{*GEODE dataset (Underground, Off-road areas) does not provide rotation ground truth.} \\

    \end{tabular}}
    \vspace{-0.6cm}
    \label{tab:slam_comparison}
\end{table*}

\subsection{Datasets}
We first selected five publicly available datasets representing aerial environments \cite{zhu2023graco}, off-road areas \cite{chen2024heterogeneous}, underground mine tunnels \cite{chen2024heterogeneous}, and custom dataset collected from our team's planetary terrains simulation open-source in NASA-JPL\footnote{\url{https://github.com/nasa-jpl/osr-rover-code.git}}~\footnote{\url{https://github.com/dongjineee/rover_gazebo.git}} for our experiments, as shown in \figref{fig:dataset}.
Additionally, we conducted experiments to check the memory usage in large-scale parks~\cite{huang2021disco} as shown in \figref{fig:main}.

\subsection{Evaluation Metrics}
\subsubsection{PR curve}
For evaluating the performance of inter-robot place recognition, the Precision-Recall (PR) curve was utilized. 
Precision and recall are defined as below:
\begin{equation}
\text { Precision }=\frac{\mathrm{TP}}{\mathrm{TP}+\mathrm{FP}}, \quad \text { Recall }=\frac{\mathrm{TP}}{\mathrm{TP}+\mathrm{FN}},
\end{equation}
where TP, FP, and FN are true positive, false positive, and false negative, respectively. 

\subsubsection{RTE, RRE, and Success Rate}
For inter-robot registration evaluations, we utilized a success rate as a key performance metric.
Thus, registration is considered successful if both translation and rotation errors are below 2$\,$m and 5$^\circ$, respectively.
Relative translation error (RTE) and relative rotation error (RRE) are utilized as follows:
\begin{itemize}
    \item $\text{RTE} = \sum^{N_{\text{success}}}_{n=1} ((t_{n,\text{GT}} - \hat t_n)^2/{N_{\text{success}}}) \times 100$,
    \item $\text{RRE} = \frac{\pi}{180} \sum^{N_{\text{success}}}_{n=1}\left\vert \cos^{-1} \bigg(\frac{\text{trace}(\hat R_n^{\texttt{T}} R_{n,\text{GT}})-1}{2}\bigg)/{N_{\text{success}}}\right\vert$,
\end{itemize}
where $t_{n,\text{GT}}$ and $\hat t_n^2$ are ground truth and estimated translation vectors, $R_{n,\text{GT}}$ and $R_n^{\texttt{T}}$  are ground truth and estimated rotation matrices, respectively; $N_{\text{success}}$ denotes the number of successful registration results.

\subsubsection{Multi-Robot Alignment Error}
To evaluate the accuracy of the estimated robot $\alpha$ with ground truth, we estimated the transformation ($\textbf{T}^{\text{GT}}_{\alpha}$) between the ground truth and the estimated pose of robot $\alpha$ using Kabsch-Umeyama algorithm~\cite{umeyama1991least}.
We then assessed the absolute pose error (APE), $\textbf{T}_{\alpha,\text{APE}}$.
Evaluating robot $\beta$ using the same method as robot $\alpha$ would solely assess trajectory accuracy, which is inadequate for a comprehensive alignment evaluation.
Therefore, to evaluate the alignment of robot $\beta$ with respect to robot $\alpha$, we utilized the alignment evaluation metric as follows:
\begin{equation}
    \textbf{T}_{\beta,\text{APE}} = (\textbf{T}^{\text{GT}}_{\beta})^{-1} \, \textbf{T}^{\text{GT}}_{\alpha} \, \textbf{T}^{\alpha}_{\beta},
\end{equation}
where $\textbf{T}^{\text{GT}}_{\beta}$ is the estimated transformation between the ground truth and the estimated pose of robot $\beta$ using \cite{umeyama1991least} and $\textbf{T}^{\alpha}_{\beta}$ is the estimated transformation of the robot $\beta$ based on the robot $\alpha$ in the back-end of distributed PGO.
The APE was derived by the EVO trajectory evaluation tool~\cite{grupp2017evo}.

\begin{figure*}[t]
	\centering
	\def\width{\textwidth}%
        {%
		\includegraphics[clip, trim= 55 900 210 380, width=\width]{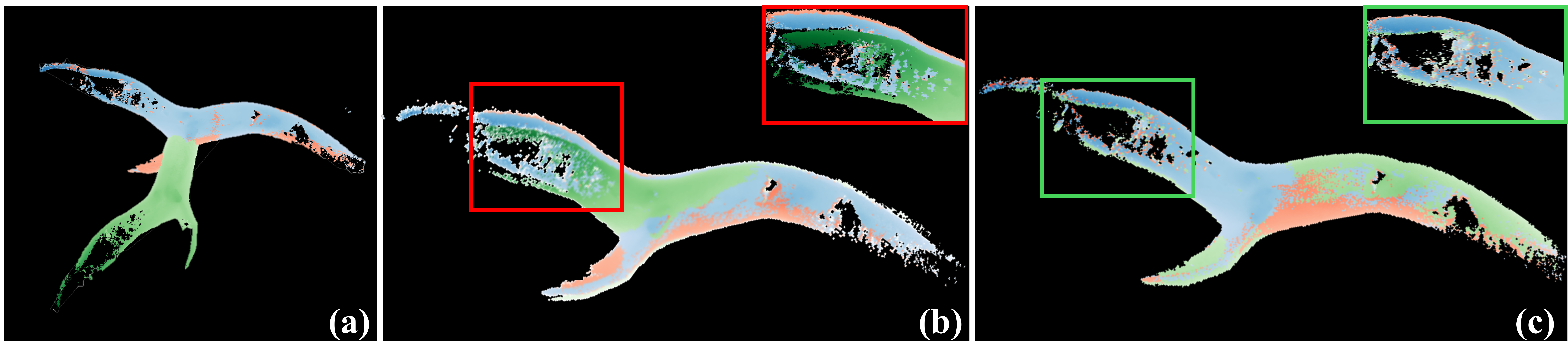}
	}
    \vspace{-0.6cm}
    \caption{Qualitative cave mapping results via distributed SLAM framework comparison: (a) DCL SLAM, (b) DiSCo SLAM, (c) Ours.}
    \label{fig:majang}
    \vspace{-0.4cm}
\end{figure*}

\subsection{Inter-robot Place Recognition}
Subsequently, inter-robot place recognition plays a crucial role in multi-robot systems by providing an initial estimate of other robots' positions.
Especially in resource-constrained field environments, lightweight global descriptors shared among robots minimize the risk of communication bottlenecks, enabling more efficient inter-robot place recognition.
\tabref{tab:desc} shows that our descriptor is 20$\times$ to 1000$\times$ lighter compared to existing distributed SLAM framework descriptors in the large-scale park dataset, collected for over 30 minutes.
Furthermore, this suggests efficiency in long-term autonomous driving scenarios, such as planetary exploration, where onboard computers operate with limited memory resources.

\figref{fig:pr_curve} illustrates the PR curves across diverse field environments.
Our method demonstrates the ability to maintain high recall in most scenarios.
This indicates that our method can capture more potential loop closures.
When combined with outlier rejection, our method can enhance overall performance, ensuring more robust and accurate mapping.

\begin{table}[b]
\vspace{-0.4cm}
\caption{Evaluation of Scan-Level Registration}
\centering\resizebox{0.48\textwidth}{!}{
\begin{tabular}{c|l|ccc}
\toprule \hline
\multicolumn{1}{c|}{\multirow{8}{*}{\rotatebox{90}{\textbf{Planetary Terrains}}}}    & \textbf{Method}
                                              & \textbf{RTE [cm] $\downarrow$}  
                                              & \textbf{RRE [$^\circ$] $\downarrow$}
                                              & \textbf{Success rate [\%] $\uparrow$} \\ \cline{2-5}
                                              & ICP \cite{rusinkiewicz2001efficient}    & 63.968                 & 2.819                          & 12.389                  \\
                                              & PLANE-ICP \cite{low2004linear}          & 47.064                 & 1.471                          & 24.447                  \\
                                              & GICP \cite{segal2009generalized}        & 7.455                  & \bbl{\textbf{0.182}}            & 71.128                  \\ 
                                              & Small-GICP \cite{small_gicp}            & 11.413                 & 0.285                          & 72.677                   \\ 
                                              & V-GICP \cite{koide2021voxelized}        & \bbl{\textbf{7.314}}    & 0.188                          & 76.770                  \\ 
                                              & KISS-Matcher \cite{lim2024kiss}         & 55.296                 & 1.460                          & 71.903                  \\ 
                                              & Ours                                    & 8.224                  & 0.281                          & \bbl{\textbf{86.394}}      \\  \hline 
\bottomrule
\end{tabular}}
\label{tab:regi_eval}
\end{table}

\subsection{Inter-robot Registration}
Inter-robot registration is an approach for estimating the relative pose between robot $\alpha$'s scan $\mathcal{P}_\alpha$ and robot $\beta$'s scan $\mathcal{P}_\beta$.
This is utilized in distributed pose graph optimization to determine the final transformation $\mathbf{T}^{\alpha}_{\beta}$ between robot $\alpha$ and $\beta$.

Our method achieves RTE and RRE values that are nearly identical to the best-performing approach, with only a $~$1$\,$cm and 1$\,$$^\circ$ gap, respectively, as shown in \tabref{tab:regi_eval}.
Since the result depends on the number of successful registrations, examining the success rate reveals that our method maintains approximately 10$\%$ greater robustness.

\begin{figure}[t]
	\centering
	\def\width{0.47\textwidth}%
        {%
		\includegraphics[clip, trim= 0 0 0 0, width=\width]{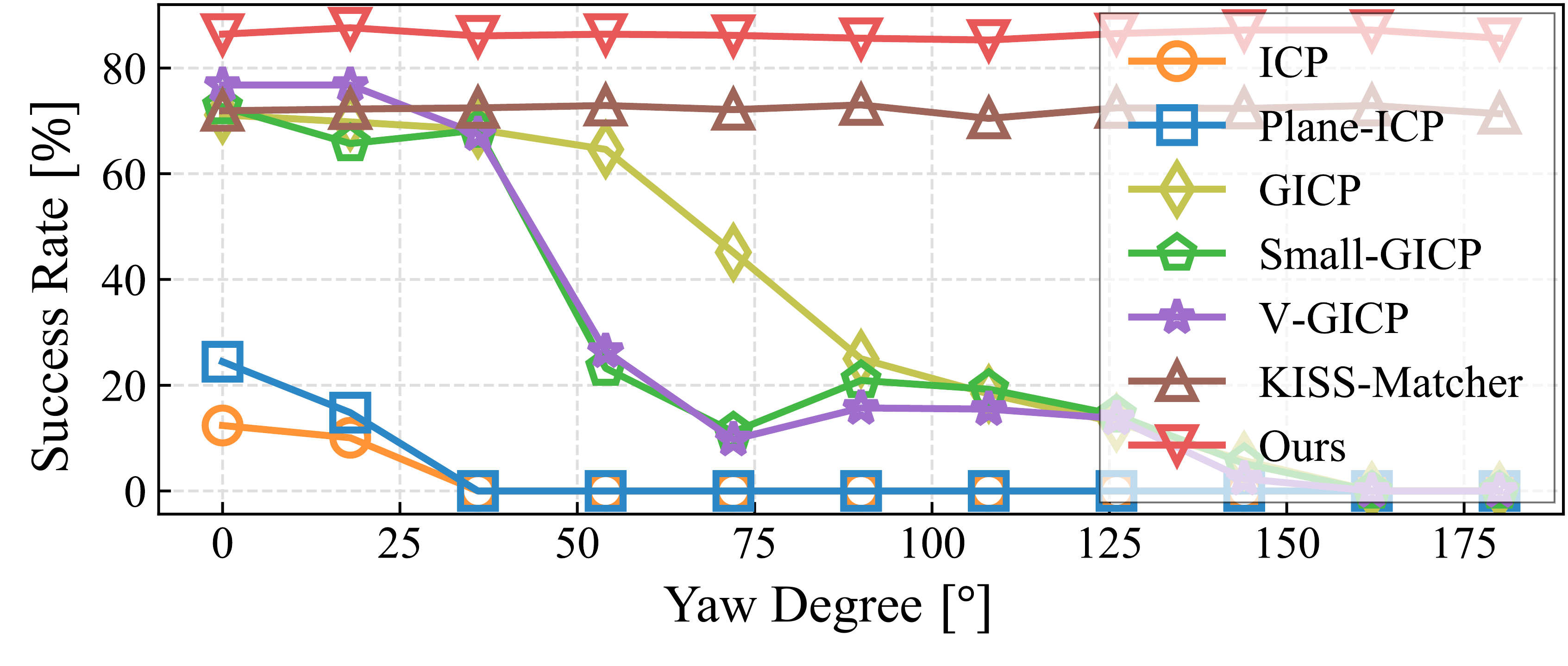}
	}
    \vspace{-0.4cm}
    \caption{The success rate when robot $\alpha$'s scan is rotated by yaw degrees.}
    \label{fig:success_rate}
    \vspace{-0.6cm}
\end{figure}

In field robotics, if robots $\alpha$ and $\beta$ navigate in different directions, a large pose discrepancy can appear between them.
Therefore, we assessed the performance invariance to pose discrepancy by rotating robot $\alpha$'s scan in nearly flat planetary terrains without tall buildings or structures as illustrated in \figref{fig:success_rate}.
Our method maintains a rotation-invariant success rate, demonstrating consistent performance regardless of the large pose discrepancy.

\subsection{Comparison with Distributed SLAM Framework}
\tabref{tab:slam_comparison} represents quantitative results of APE, separated into absolute rotation error (ARE) and absolute translation error (ATE) components across various field environments.
Our framework exhibits low error, maintaining field-agnostic performance.
Especially, our framework performs robust alignment even in aerial environments, where partial overlap occurs due to robots $\alpha$ and $\beta$ traveling in completely inverse directions as shown in \figref{fig:trajectories}.
On the other hand, in featureless off-road areas, alignment was achieved but lacked high precision.

\begin{figure}[b!]
    \vspace{-0.5cm}
    \centering
	\def\width{0.48\textwidth}%
        {%
		\includegraphics[clip, trim= 40 0 0 0, width=\width]{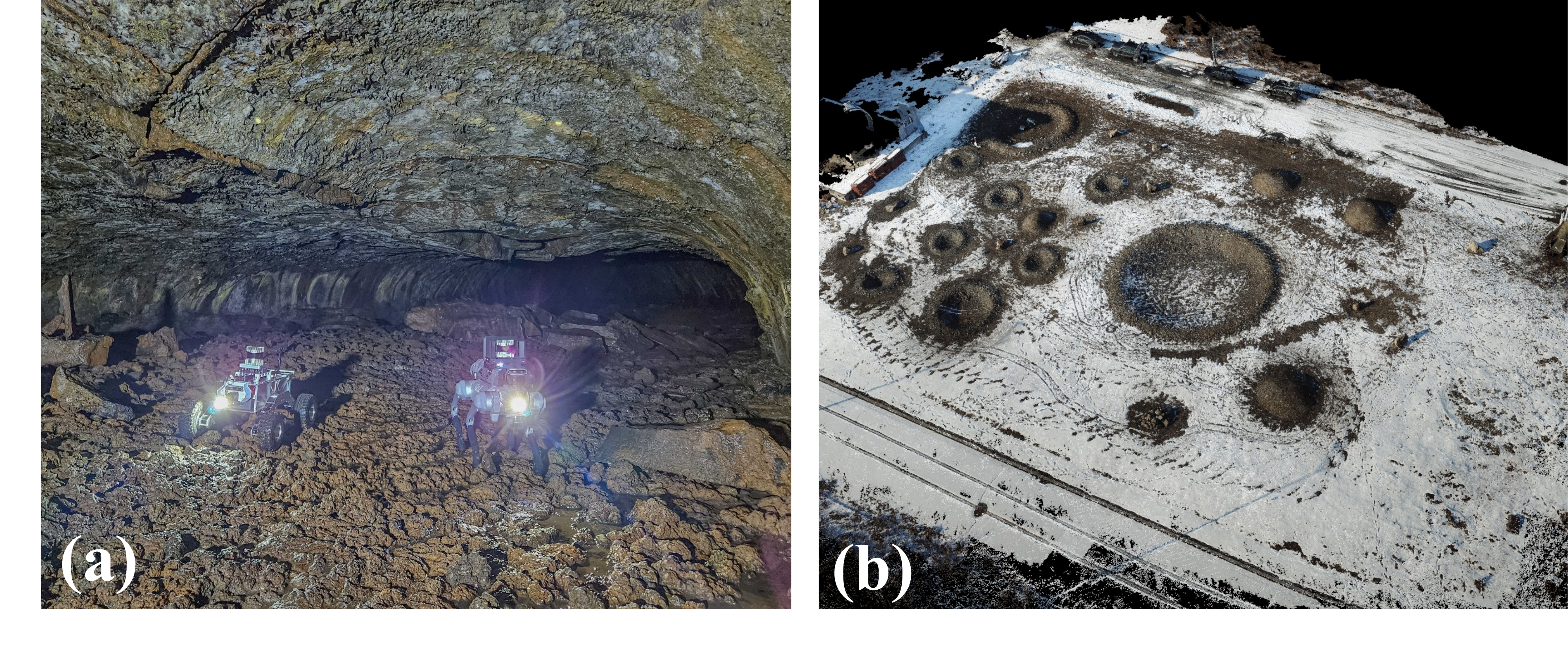}
	}
    \vspace{-0.8cm}
    \caption{(a) Caves of our in-house field datasets. (b) Planetary terrains (of drone view) in-house field datasets. 
              Note that the planetary emulation environment refers to a real-world outdoor terrain designed to resemble extraterrestrial surfaces, and is not a simulated or virtual environment.
}
    \label{fig:custom_dataset}
\end{figure}

%% file: PaperWriting/field_experiments.tex
\section{Field Evaluation}
To demonstrate the field versatility of our method, as shown in Section \uppercase\expandafter{\romannumeral5}, we evaluated it on our planetary emulation terrains and caves environment as shown in \figref{fig:custom_dataset}. 
Unfortunately, obtaining ground truth in such field environments is challenging. 
Therefore, we demonstrate the robustness of our method through qualitative results.

\subsection{Field setup}
We deployed a quadruped robot (Unitree GO-2) and a custom rover, as shown in \figref{fig:custom_dataset} (a), to collect data from planetary emulation terrains and a cave, respectively.
The quadruped robot was equipped with an Ouster-32 LiDAR and a MicroStrain CV7 IMU, and the rover was configured with an Ouster-64 LiDAR and a MicroStrain GX5-25 IMU.

\subsection{Communication and Memory}
For evaluating the exchange of global descriptors, we deployed an Omada Wireless WiFi AP for robot communication, utilizing ROS messages and ZeroMQ \cite{hintjens2013zeromq}.
We measured latency by recording the time when robots $\beta$ and $\gamma$ sent the global descriptor and the time when robot $\alpha$ received the corresponding data as represented in \tabref{tab:latency}.
Our method maintains a consistently low latency of 0.05 to 0.06 seconds, regardless of the distance between robots, compared to other approaches.
Notably, LiDAR Iris descriptor experiences bottlenecks starting at 20$\,$m, where the WiFi signal weakens.

\begin{figure}[t]
	\centering
	\def\width{0.48\textwidth}%
        {%
		\includegraphics[clip, trim= 0 0 0 0, width=\width]{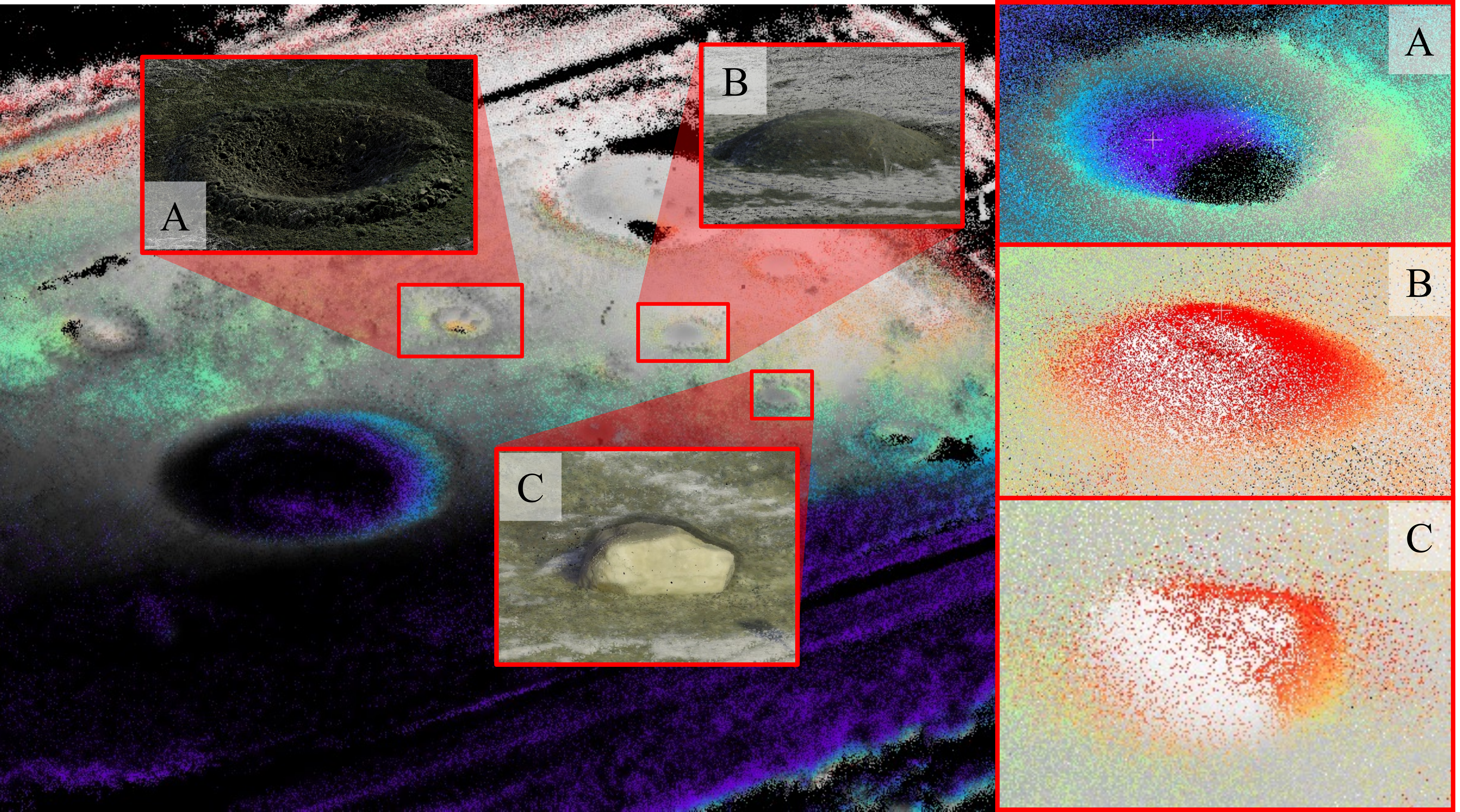}
	}
    \vspace{-0.7cm}
    \caption{Multi-robot mapping of the planetary terrains using our algorithm.}
    \label{fig:steam}
    \vspace{-0.1cm}
\end{figure}

\begin{table}[t]
\caption{Latency Comparison}
\renewcommand{\arraystretch}{1.2}
\centering\resizebox{0.48\textwidth}{!}{\tiny
\begin{tabular}{ll|ccc|ccc}
\toprule \hline
\multicolumn{2}{l|}{\textbf{Datasets}} & \multicolumn{3}{c|}{\textbf{Majang (Sharing: $\beta, \ \gamma \rightarrow \alpha$)}} & \multicolumn{3}{c}{\textbf{Planetary Emulation (Sharing: $\beta \rightarrow \alpha$)}} \\ \hline
\multicolumn{2}{l|}{\textbf{Desc.}}                         & SC [sec]  & IRIS [sec]  & Ours [sec]            & SC [sec]  & IRIS [sec]    & Ours [sec]          \\ \hline
\multicolumn{1}{l|}{\multirow{6}{*}{\textbf{\rotatebox{90}{Meesage Time}}}} 
                                                & 5\,m        & 0.102     & 68.552      & \bbl{\textbf{0.034}}   & 0.070     & 55.392        & \bbl{\textbf{0.030}} \\ 
\multicolumn{1}{l|}{}                           & 10\,m       & 0.103     & 84.532      & \bbl{\textbf{0.035}}   & 0.069     & 70.907        & \bbl{\textbf{0.032}}  \\
\multicolumn{1}{l|}{}                           & 15\,m       & 0.160     & 281.914     & \bbl{\textbf{0.038}}   & 0.120     & 230.129       & \bbl{\textbf{0.036}}  \\
\multicolumn{1}{l|}{}                           & 20\,m       & 0.674     & $\geq$45m   & \bbl{\textbf{0.041}}   & 0.581     & $\geq$30m     & \bbl{\textbf{0.039}}   \\
\multicolumn{1}{l|}{}                           & 25\,m       & 1.151     & $\geq$1h    & \bbl{\textbf{0.046}}   & 0.945     & $\geq$1h      & \bbl{\textbf{0.040}}   \\
\multicolumn{1}{l|}{}                           & 30\,m       & 1.523     & $\geq$2h    & \bbl{\textbf{0.060}}   & 1.321     & $\geq$1h 30m  & \bbl{\textbf{0.054}}   \\ \hline \bottomrule
\end{tabular}}
\label{tab:latency}
\vspace{-0.6cm}
\end{table}

\subsection{Distributed LiDAR Mapping}
\figref{fig:majang} and \figref{fig:steam} show the mapping results of various distributed LiDAR SLAM frameworks, including our method, in the cave and the mapping results of our method in planetary emulation terrains, respectively.
Our method generated a consistent map in the cave without distortions.
In contrast, in DCL SLAM, LiDAR Iris failed to perform descriptor matching, while in DiSCo SLAM, ICP-based relative pose estimation was not reliable.
In the planetary emulation terrains, when RGB images were overlaid, the craters, ridges, and rock contours were clearly defined, indicating that the SKiD-SLAM was successfully performed.

%% file: PaperWriting/conclusion.tex
\section{Conclusion}
In this study, we propose a novel distributed LiDAR SLAM framework, SKiD-SLAM that can be applied to versatile field environments.
We demonstrate that our framework maintains robustness in field environments.
Also, real-world communication experiments confirm that our method enables fast data transmission without noticeable latency.
In future works, we plan to extend our SLAM pipeline to $N\geq4$ robot systems.